\documentclass[letterpaper, 10 pt, conference]{ieeeconf}  

\IEEEoverridecommandlockouts                              

\usepackage{graphics} 
\usepackage{epsfig} 
\usepackage{times} 
\usepackage{amsmath} 
\usepackage{amssymb}  

\usepackage{booktabs}
\usepackage{siunitx}
\usepackage{booktabs} 
\usepackage{amsmath,amssymb}
\usepackage{siunitx}      
\title{\LARGE \bf
Designing an Efficient Excavator Bucket for Lunar ISRU: A Comparative Study with Vision-Based Fill and Displacement Analysis
}

\author{Abdulla Hil Kafi$^{1}$, Tomoki Koshi$^{1}$, Jorge Casir$^{1}$ and Kenji Nagaoka$^{1}$
\thanks{$^{1}$Abdulla Hil Kafi, Tomoki Koshi, Jorge Casir, and Kenji Nagaoka are with Department of Mechanical and control Engineering, Kyushu Institute of Technology,
        Kitakyushu 804-8550, Japan
       {\tt\small kafi-abdulla-hil202@mail.kyutech.jp};
       {\tt\small nagaoka.kenji572@mail.kyutech.jp}}%
}

\begin{document}

\maketitle
\thispagestyle{empty}
\pagestyle{empty}

\begin{abstract}

This paper present a spiral-cavity wheel for lunar regolith excavation and a sensor-light evaluation stack that jointly estimates fill ratio (vision), sinkage (vision), and specific energy from actuator logs. In benchtop tests (four revolutions at 5, 10, and 15~RPM) against two literature baselines, the proposed wheel achieved higher excavated mass and fill ratio, delivering 2.2--3.0 times higher excavation rate while reducing specific energy by 29\% relative to a bucket-drum baseline. Normalized sinkage (mm/kg) was also lower, indicating stable traction without bogging. Effort-time traces show a steady torque envelope with repeatable cut--carry--dump cycles across speeds. We provide a retention index $\eta$ that correlates with fill ratio and a DEM setup that reproduces experimental trends with low error. Results suggest spiral-cavity wheels can replace heavier multi-actuator diggers when mass, simplicity, and energy efficiency are mission drivers.

\end{abstract}

\section{INTRODUCTION}

In-situ resource utilization (ISRU) is central to sustaining long-term lunar missions by reducing reliance on Earth-supplied resources~\cite{c9}. Excavation of regolith is particularly important, providing feedstock for oxygen production, shielding, and construction. Yet lunar excavation faces unique challenges: reduced gravity alters soil–tool dynamics, regolith particles are sharp and abrasive, and mission constraints demand systems that are lightweight and energy-efficient.  

Conventional methods such as bucket drums and robotic arms have been studied extensively. Bucket drums achieve high efficiency~\cite{c1} but require deployment mechanisms that add mass and complexity, while robotic arms impose high power and control demands. These limitations motivate simpler, integrated solutions.  

Planetary rover wheels, widely studied for traction in granular soils~\cite{c2,c25,c26}, remain underexplored as excavation tools. A wheel capable of both mobility and regolith collection could reduce subsystem redundancy, provided its performance is experimentally validated. Key trade-offs between efficiency, energy use, and sinkage must therefore be quantified.

This paper presents the design and experimental evaluation of a spiral-cavity excavator wheel for regolith collection. The wheel was tested at multiple rotational speeds in a sandbox environment and benchmarked against two baseline designs. To enable quantitative analysis, we developed lightweight vision- and log-based tools for fill ratio, sinkage, and excavation-specific energy.  
\begin{figure}[t]
      \centering
      \includegraphics[width=0.31\linewidth]{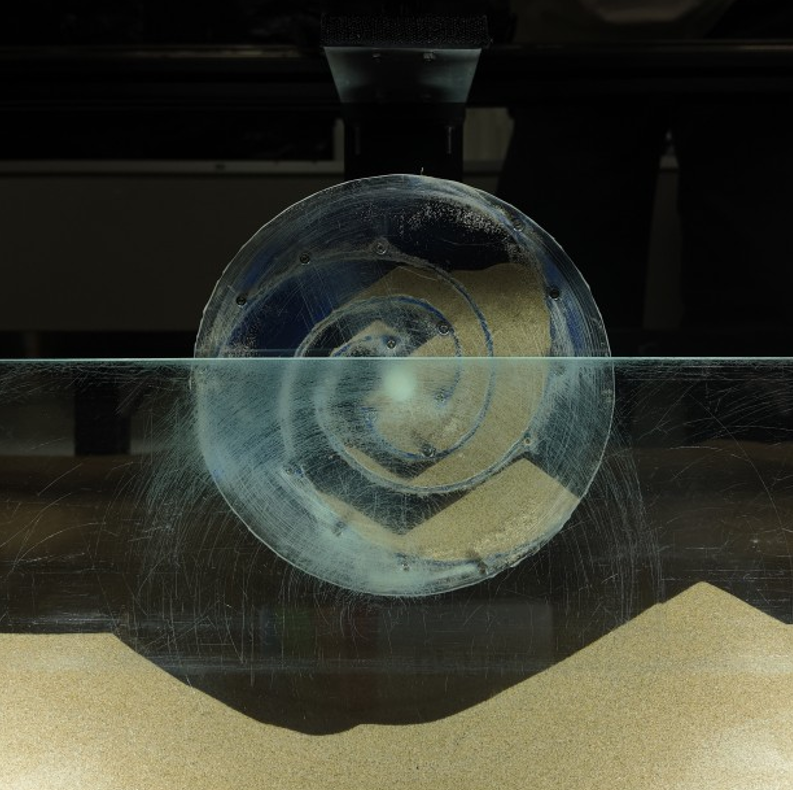}\hfill
      \includegraphics[width=0.31\linewidth]{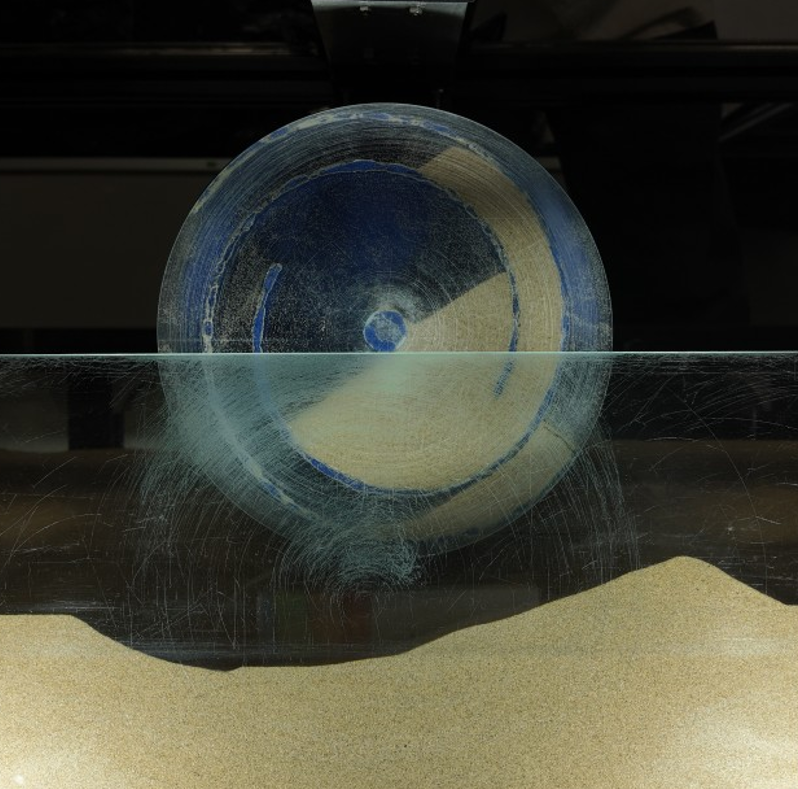}\hfill
      \includegraphics[width=0.31\linewidth]{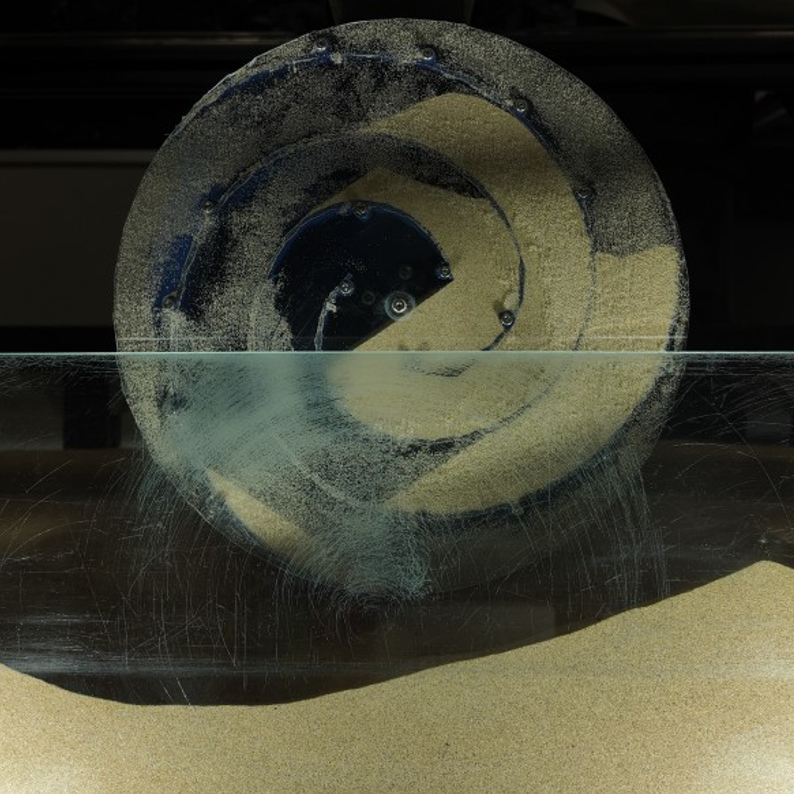}
      \caption{Wheels under test: two baselines (left, middle) and the proposed spiral-cavity design (right).}
      \label{fig:Thre_Design}
\end{figure}

The main contributions are:
\begin{itemize}
    \item A novel spiral-cavity wheel design optimized for regolith retention.
    \item Experimental benchmarking against two literature-derived baselines.
    \item Vision- and log-based tools for excavation metrics without external instrumentation.
    \item Analysis of excavation-specific energy and trade-offs in efficiency.
\end{itemize}

\section{Motivation}
Previous studies have investigated four-wheeled lunar rover platforms equipped with active suspension systems and one-way clutch mechanisms to enhance mobility resilience under actuator failure and challenging terrain conditions~\cite{c2}. These designs demonstrated fault-tolerant locomotion and the capability to traverse soft regolith, highlighting that lightweight mechanical architectures can reduce the need for redundant subsystems while maintaining operational reliability. Building on these insights, the present work extends the rover’s functionality toward in-situ resource utilization (ISRU). Instead of incorporating a dedicated excavation mechanism, this study proposes a wheel design capable of simultaneously supporting locomotion and regolith collection. Such integration aims to reduce overall system mass, power consumption, and mechanical complexity while enabling excavation tasks relevant to ISRU and lunar surface construction.  

\begin{figure}[t]
      \centering
      \includegraphics[width=.95\linewidth]{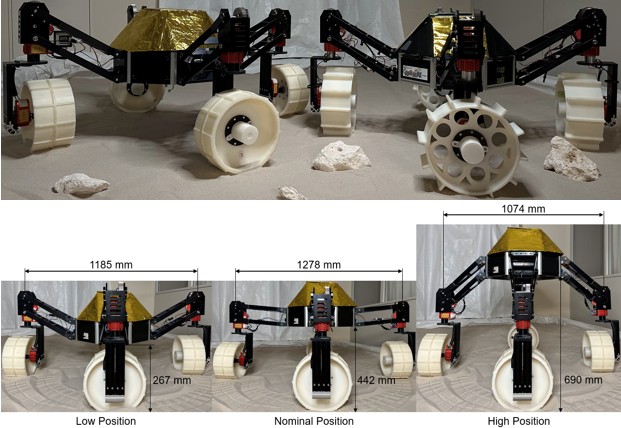}
      \caption{Lunar Rover Prototypes and Configuration}
      \label{fig:rover}
\end{figure}

\section{Related Work}

Excavation of lunar regolith has traditionally relied on bucket drum systems. Li et al.~\cite{c1} showed that drum geometry and scoop spacing strongly influence efficiency, and NASA’s RASSOR prototypes use counter-rotating drums to minimize reaction forces~\cite{c10,c21}. While effective, these designs require deployment mechanisms that add mass and actuator complexity, motivating interest in more integrated excavation solutions.  

Planetary wheels have been widely studied in terramechanics. Kamohara et al.~\cite{c8} modeled grousered wheels in simulants, and Keio University~\cite{c7} examined drum-type wheels with discrete scoops for retention. Recent single-wheel tests, such as Takehana~\cite{c14} on cohesive soils, underline how geometry and soil properties govern traction, torque, and sinkage. However, excavation is typically treated as incidental to locomotion rather than a design objective.  

Discrete Element Method (DEM) modeling is commonly applied to simulate granular flow and excavation forces. Properly calibrated, DEM can reproduce experimental trends and guide system design~\cite{C11,c20}. Abubakar et al.~\cite{c18} modeled Rashid rover wheels with partial validation, yet DEM remains under-validated for wheel excavators across geometries and operating conditions.  

Experimental work has traditionally relied on load cells or motion capture. More recently, vision-based methods estimate slip and sinkage in real time~\cite{C12}, and current-based models have been used for trafficability analysis~\cite{c16}. Still, these approaches are seldom applied to excavation-specific metrics like fill ratio or penetration depth, and excavation-specific energy is rarely quantified despite its relevance to ISRU feasibility.  

At the system level, excavation is increasingly framed as part of lunar infrastructure. Xia et al.~\cite{c19} identify regolith handling as a construction bottleneck, and Luna et al.~\cite{c17} report field results from modular rover campaigns, linking mobility and excavation to Artemis-class mission feasibility. Against this background, our work introduces a wheel design that integrates excavation into rover architecture and contributes lightweight, vision-based tools for performance evaluation without external instrumentation.

\section{Wheel Designs}
Three different wheel designs were evaluated to investigate excavation performance under controlled conditions.
\subsection{Design A (Proposed Wheel)}

Design A is the novel wheel proposed in this study and is intended for integration with a four-wheeled rover platform previously reported in the literature \cite{c2}, \cite{c3}. That rover architecture was primarily designed to ensure resilient mobility and fault-tolerant operation in challenging terrain environments, but it did not include a dedicated excavation system for regolith collection. To extend this capability toward in-situ resource utilization (ISRU), we designed a new wheel that could be directly integrated with the rover’s drivetrain while enabling regolith collection.
\begin{figure}[t]
      \centering
      \includegraphics[width=.75\linewidth]{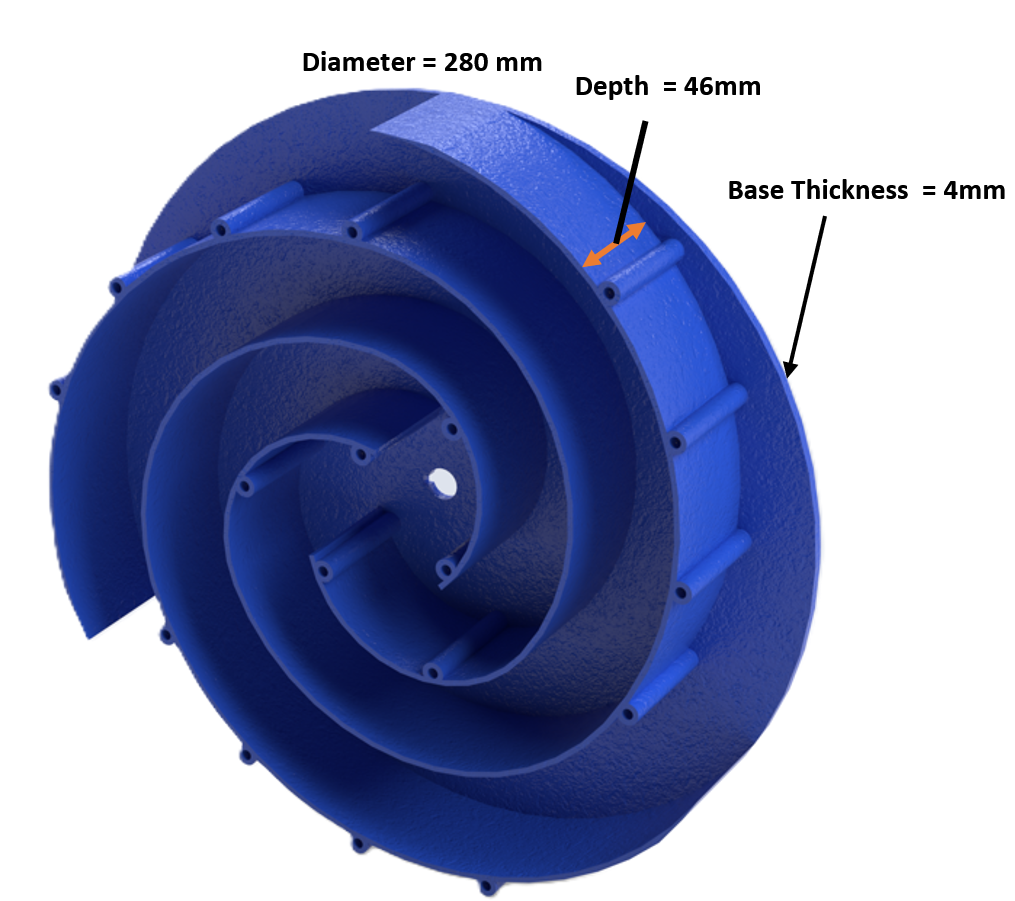}
      \caption{CAD model of the bucket-drum excavator showing the scoop geometry used for 3D printing}
      \label{fig:3D}
\end{figure} 
The wheel has a diameter of approximately 0.280~m and a depth of 0.046~m, dimensions selected to match the rover’s suspension geometry and actuator torque limits \cite{c2} , \cite{c3}. A distinguishing feature of Design A is the central spiral cavity, intended to increase soil retention during excavation. Unlike Designs B and C, which feature open centers where material can readily spill out, the spiral cavity of Design A acts as a passive regolith holder, reducing losses during lifting. The cavity was generated by connecting spline curves under tangential continuity (G$^1$ constraint), ensuring smooth transitions without sharp edges that might destabilize soil flow. This geometry encourages inward soil guidance and retention throughout the excavation cycle. The spiral geometry followed specific design considerations:
The spiral geometry was defined by the following criteria: i) the spiral completes one full 360$^\circ$ turn around the axis of the wheel; ii) the radial distance from the inner scoop end to the wheel center was fixed at 0.2 times the wheel radius; iii) scoops were positioned at uniform angular intervals to maintain consistent spacing and geometric alignment; and iv) a tangential constraint was applied when connecting adjacent curves, producing a continuous spiral profile for smooth soil flow.

To formalize this feature, we introduce a \textit{soil-holding ratio} ($\eta$), which provides a dimensionless measure of the wheel’s internal capacity to retain regolith:
\begin{equation}
    \eta = \frac{V_{\text{cavity}}}{V_{\text{wheel}}}
\end{equation}
where $V_{\text{cavity}}$ is the approximate cavity volume and $V_{\text{wheel}}$ is the bounding cylindrical volume of the wheel. For practical estimation, both quantities can be approximated from external dimensions (outer diameter $D$, wheel width $W$, and cavity depth $h_c$) or simplified to a 2D cross-sectional ratio between cavity area and wheel area. This avoids dependence on detailed CAD integration while still capturing the relative differences between designs. The soil-holding ratio is not intended as an exact volumetric measurement, but rather as a comparative index of retention potential across wheel geometries. In the Results section, we show that the ranking of $\eta$ correlates with experimentally measured fill ratios, supporting its validity as a predictive design metric.

The final geometry was modeled in CAD and fabricated using a fused deposition modeling (FDM) 3D printer with PLA filament. This rapid prototyping approach enabled iterative refinement while maintaining sufficient structural strength for laboratory-scale experiments.  Figure~\ref{fig:3D} illustrates the CAD model of Design A, highlighting the spiral cavity and construction features.

\subsection{Design B (Bucket Drum-Inspired Wheel):} Design B was based on a bucket drum concept previously reported for lunar regolith excavation ~\cite{c1}. The geometry consists of uniformly spaced buckets distributed around the wheel periphery, which provide strong theoretical excavation capacity but also impose higher torque demand due to aggressive cutting profiles. The CAD model was reconstructed from published dimensions to ensure comparability with Design A. A PLA prototype was fabricated via FDM printing for laboratory evaluation. 

\subsection{Design C (Keio Wheel):} Design C replicates a wheel originally proposed in excavation studies conducted at Keio University ~\cite{c7}. Unlike the bucket drum, this design incorporates shallower scoops and surface features intended to improve retention while keeping torque requirements moderate. It represents a hybrid approach between soil cutting and regolith lifting. A PLA prototype was produced from reconstructed CAD models to match the scale of Design A and B for fair comparison. Together, these three designs provide a comparative study between a new excavation-optimized wheel (Design A), a bucket drum-inspired wheel (Design B), and a hybrid geometry from existing literature (Design C).
\begin{table}[h]
    \centering
    \caption{Wheel Geometry Used in Experiments}
    \label{tab:geom}
    \begin{tabular}{lcc}
        \toprule
        \textbf{Wheel} & \textbf{Inner Diameter $D$ (m)} & \textbf{Depth $L$ (m)} \\
        \midrule
        Design A (Proposed) & 0.280 & 0.0460 \\
        Design B (Baseline) & 0.229 & 0.0336 \\
        Design C (Baseline) & 0.200 & 0.0460 \\
        \bottomrule
    \end{tabular}
\end{table}

\section{Experiment Setup}
\subsection{Testbed-based excavation experiment}
\begin{figure}[t]
      \centering
      \includegraphics[width=\linewidth]{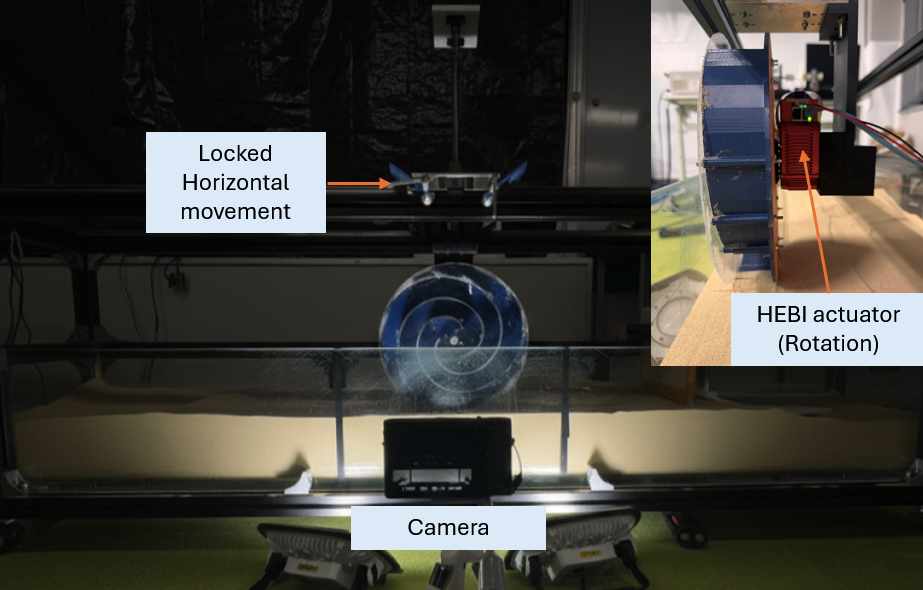}
      \caption{Experimental platform for quantitative assessment of wheel-soil excavation performance }
      \label{fig:exca}
\end{figure}
The experiments were performed using a custom-built laboratory testbed. Each wheel prototype was mounted on a rigid shaft and driven by a HEBI X8-16 smart actuator. This motor was selected due to its integrated sensors, which provide real-time torque, angular velocity, and energy consumption data. Such integrated measurements eliminated the need for external load cells and allowed direct comparison across wheel designs.This actuator is also used in our rover for steering, suspension, and mobility. The wheels were tested in a fixed position, partially submerged in a container filled with silica sand with bulk density ~1282 kg/m³ and mean particle size ~510 µm. Prior to each test, the surface of the simulant bed was leveled to maintain consistent initial conditions.

Each wheel design was tested at three rotational speeds: 5, 10, and 15 RPM. For each trial, the wheel was rotated continuously to excavation cycles of 48, 24, and 12 s. After excavation, the collected silica sand was carefully removed and weighed to determine the mass collection. Every test condition was repeated three times, and results were checked to improve reliability and reduce the effect of random variability.

This experimental protocol enabled a controlled comparison of excavation efficiency, torque demand, and energy consumption across the three wheel geometries.

\subsection{DEM Simulation}
\begin{figure}[thpb]
      \centering
      \includegraphics[width=\linewidth]{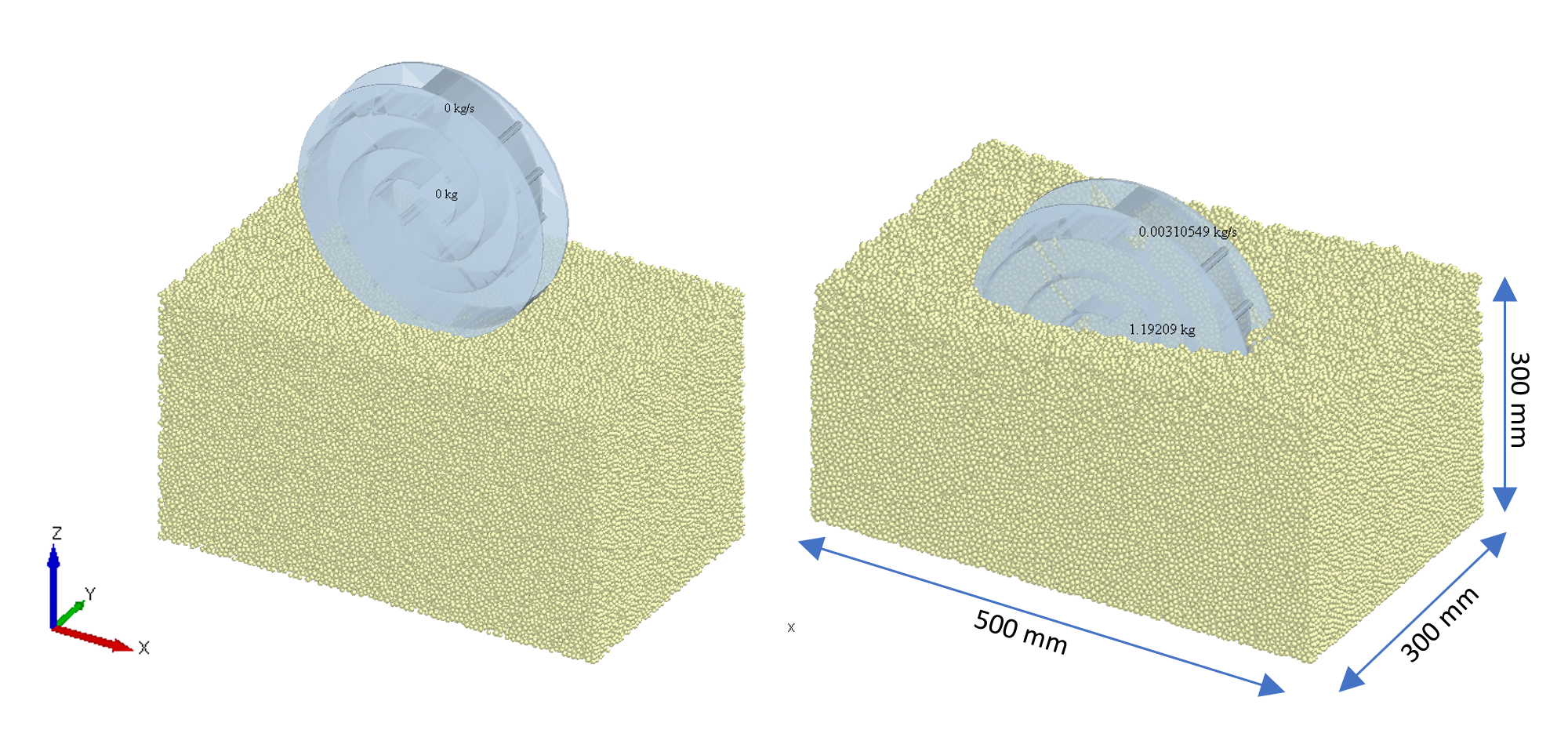}
      \caption{DEM simulation setup and excavation outcome using Altair EDEM. .}
      \label{dem_setup}
\end{figure}
The DEM simulations, conducted using Altair EDEM, showed strong agreement with experimental findings.In Fig.~\ref{dem_setup} , the left panel shows the initial state of the cylindrical tool embedded in a granular block (500~mm~$\times$~300~mm~$\times$~300~mm), while the right panel illustrates the post-excavation state with total excavated mass (1.192~kg). Predicted mass trends closely matched observed excavation outcomes, reinforcing confidence in the simulation setup. All simulation parameters were adopted from Nakano et al.~\cite{c7}, with the exception of bulk density, which was set to 1282 kg/m³ to reflect the specific properties of the experimental material. 

\section{Data Acquisition and Analysis}

\subsection{Vision-Based Fill Ratio and Mass}
\begin{figure}[t]
      \centering
      \includegraphics[width=\linewidth]{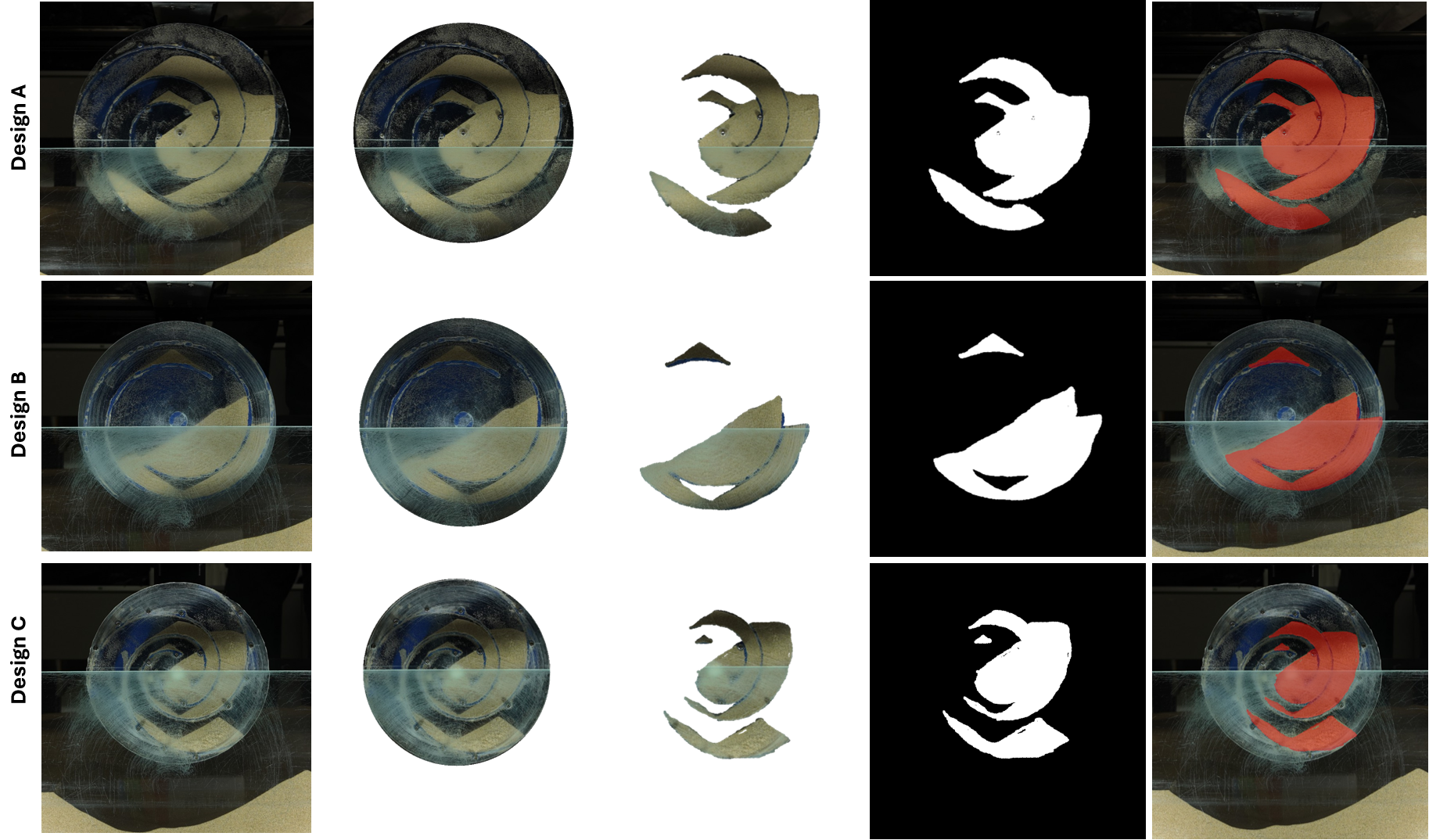}
      \caption{Fill Ratio Analysis of Wheel at 15 RPM using Image Segmentation}
      \label{fig:fillrat}
\end{figure}
A circular region of interest (ROI) is obtained by either Hough detection or one-click center and rim. Let $C\subset\mathbb{Z}^2$ be pixels inside the wheel circle of center $(c_x,c_y)$ and radius $r$. A binary mask $S(x,y)\in\{0,1\}$ segments regolith within $C$ (HSV thresholding, $k$-means in CIE–Lab, or interactive GrabCut).
\begin{align}
P_t &= \sum_{(x,y)\in C} S(x,y), &
A &= \sum_{(x,y)\in C} 1, \label{eq:pixel_counts}
\end{align}
where $P_t$ is the regolith pixel count and $A$ is the circle area in pixels. The simple fraction is $f=P_t/A$. With optional two-point calibration (``empty'' $P_0$ and ``full'' $P_{\max}$),
\begin{equation}
R_t=\mathrm{clip}\!\left(\frac{P_t-P_0}{P_{\max}-P_0},\,0,\,1\right),
\label{eq:fill_cal}
\end{equation}
which we call the \emph{calibrated fill ratio}. Given inner diameter $D$, axial depth $L$, and bulk density $\rho$,
\begin{equation}
V = R_t\,\pi\!\left(\tfrac{D}{2}\right)^{\!2} L, 
\qquad 
m=\rho\,V ,
\label{eq:mass_from_fill}
\end{equation}
yielding an image-based mass estimate $m$.

\subsection{Video-Based Vertical Displacement (Sinkage)}

\begin{figure}[t]
      \centering
      \includegraphics[width=.7\linewidth]{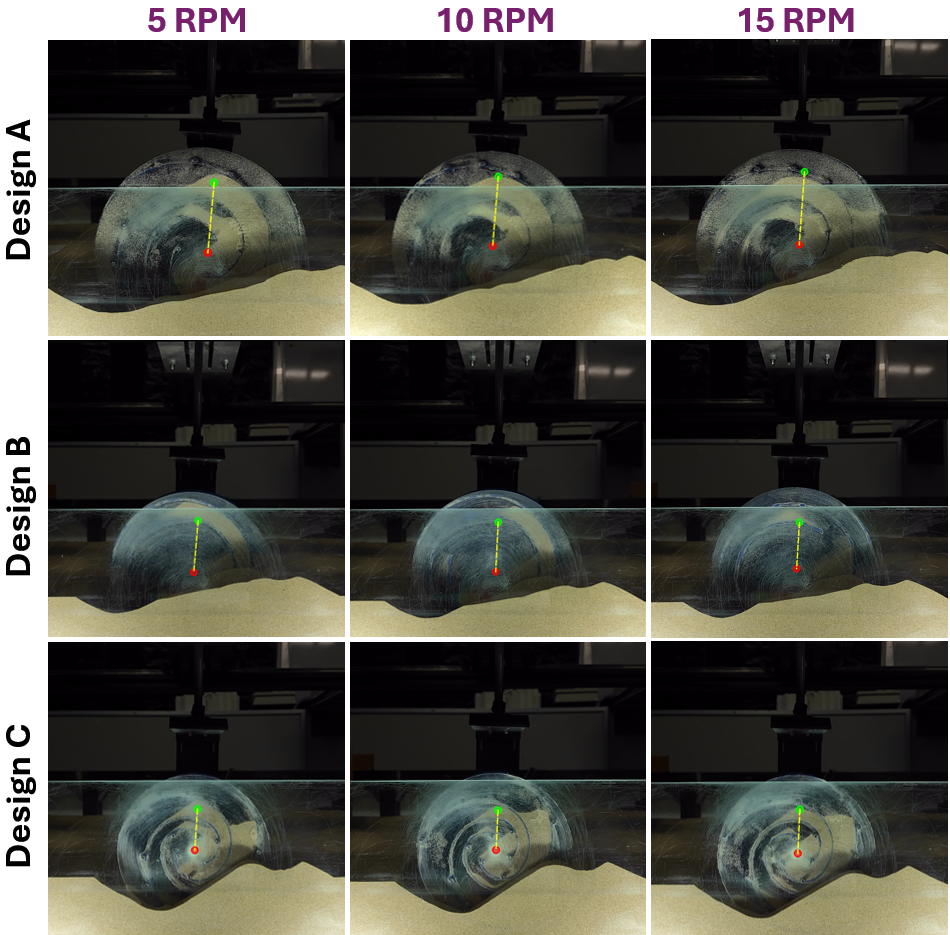}
      \caption{Displacement analysis of the excavator wheel using the custom tracking tool. The green marker represents the wheel center before excavation, and the red marker denotes the center after excavation, both derived from rim fitting. Pixel-to-millimeter calibration was performed using the known wheel diameter within the software interface. }
      \label{fig:sinkage}
\end{figure}
A pixel-to-metric scale $\kappa$ [mm/pixel] is set by dragging across a known diameter $D_{\mathrm{mm}}$ that measures $D_{\mathrm{px}}$ pixels in the frame:
\begin{equation}
\kappa=\frac{D_{\mathrm{mm}}}{D_{\mathrm{px}}}.
\end{equation}
Start and end positions $(x_0,y_0)$, $(x_1,y_1)$ are taken from user clicks or from algebraic circle fits to rim points (start/end frames). Displacements are
\begin{equation}
\begin{aligned}
\Delta x &= (x_1 - x_0)\,\kappa,\\
\Delta y &= (y_1 - y_0)\,\kappa,\\
\Delta s &= \sqrt{\Delta x^{2} + \Delta y^{2}} .
\end{aligned}
\label{eq:disp}
\end{equation}
We report sinkage as the vertical component $\Delta y$ (mm).


\subsection{Actuator-Log Power, Energy, and Rate}
From actuator effort $\tau(t)$ [\si{\newton\metre}] and angular velocity $\omega(t)$ [\si{\radian\per\second}],
\begin{equation}
P_m(t)=\tau(t)\,\omega(t), \qquad 
E_m=\int_{t_0}^{t_1} P_m(t)\,dt ,
\label{eq:power_energy}
\end{equation}
using trapezoidal integration over the trimmed interval $[t_0,t_1]$.
The \emph{specific energy} is $E_m/m$ and the \emph{excavation rate} is $m/(t_1-t_0)$.

When position $\theta(t)$ is available, we unwrap to $\tilde{\theta}(t)$ and segment by revolutions to obtain
\begin{equation}
E_k=\int_{2\pi(k-1)}^{2\pi k}\!\tau(\theta)\,d\theta, \qquad 
E_m=\sum_k E_k ,
\label{eq:per_rev}
\end{equation}
which also yields torque–angle loops $\tau(\theta)$ showing cut–carry–dump cycles.



\section{Results and Discussion}
The following results are based on excavation trials conducted with silica sand as the regolith simulant. While silica sand differs from lunar regolith in mineral composition and particle angularity, it remains a reliable analogue for capturing wheel--soil interaction trends and for benchmarking DEM predictions. Table~\ref{tab:results} summarizes the nine test conditions (three wheel designs $\times$ three rotational speeds).

\begin{table*}[t]
\centering
\caption{Comparative results for 4 revolutions per test. A: proposed; B,C: baselines. Excavation rate = collected mass / test time (48, 24, 16 s at 5/10/15 RPM).}
\label{tab:results}
\small
\setlength{\tabcolsep}{2.8pt}      
\renewcommand{\arraystretch}{1.1}
\begin{tabular}{@{}l c c c c c c c c c@{}}
\toprule
\textbf{Wheel} & \textbf{RPM} & \textbf{Fill (\%)} & \textbf{Mass (kg)} &
\textbf{Rate (kg/s)} & \textbf{Energy (J)} & \textbf{Spec.\ E (J/kg)} &
\textbf{Sink. (mm)} & $\boldsymbol{\bar{T}}$ \textbf{(Nm)} & $\boldsymbol{T_{\max}}$ \textbf{(Nm)} \\
\midrule
A & 5  & 34.6 & 1.322 & 0.027 & 57.07 & 43.17 & 93.4  & 4.10 & 6.28 \\
A & 10 & 38.2 & 1.366 & 0.056 & 73.16 & 53.56 & 99.8  & 3.55 & 5.56 \\
A & 15 & 39.2 & 1.424 & 0.086 & 56.59 & 80.59 & 105.5 & 3.38 & 5.34 \\
\midrule
B & 5  & 35.6 & 0.626 & 0.013 & 55.73 & 89.03 & 74.1 & 2.23 & 3.66 \\
B & 10 & 32.9 & 0.570 & 0.024 & 43.15 & 75.70 & 69.7 & 1.70 & 3.53 \\
B & 15 & 28.7 & 0.490 & 0.029 & 42.06 & 85.83 & 65.1 & 1.75 & 3.48 \\
\midrule
C & 5  & 28.1 & 0.514 & 0.011 & 24.84 & 48.33 & 54.5 & 0.99 & 2.46 \\
C & 10 & 30.1 & 0.521 & 0.021 & 22.79 & 43.74 & 51.2 & 0.89 & 2.45 \\
C & 15 & 28.9 & 0.512 & 0.031 & 20.71 & 40.44 & 56.4 & 0.87 & 2.41 \\
\bottomrule
\end{tabular}
\normalsize
\end{table*}

\subsection{Excavation Capacity}
The proposed wheel (Design A) consistently outperformed Designs B and C in terms of both fill ratio and excavation rate. Averaged across all speeds, Design A achieved a mean fill ratio of 37.3\%, compared with 32.4\% for Design B and 29.0\% for Design C. This corresponds to relative improvements of approximately 15\% over Design B and 29\% over Design C. At 5~RPM, Design A reached a fill ratio of 34.6\%, while Designs B and C recorded 35.6\% and 30.5\%, respectively. At higher speeds (10 and 15~RPM), Design A maintained superior performance, with fill ratios of 38.2\% and 39.2\%, whereas Designs B and C did not exceed 33\%.  
\begin{figure}[t]
      \centering
      \includegraphics[width=\linewidth]{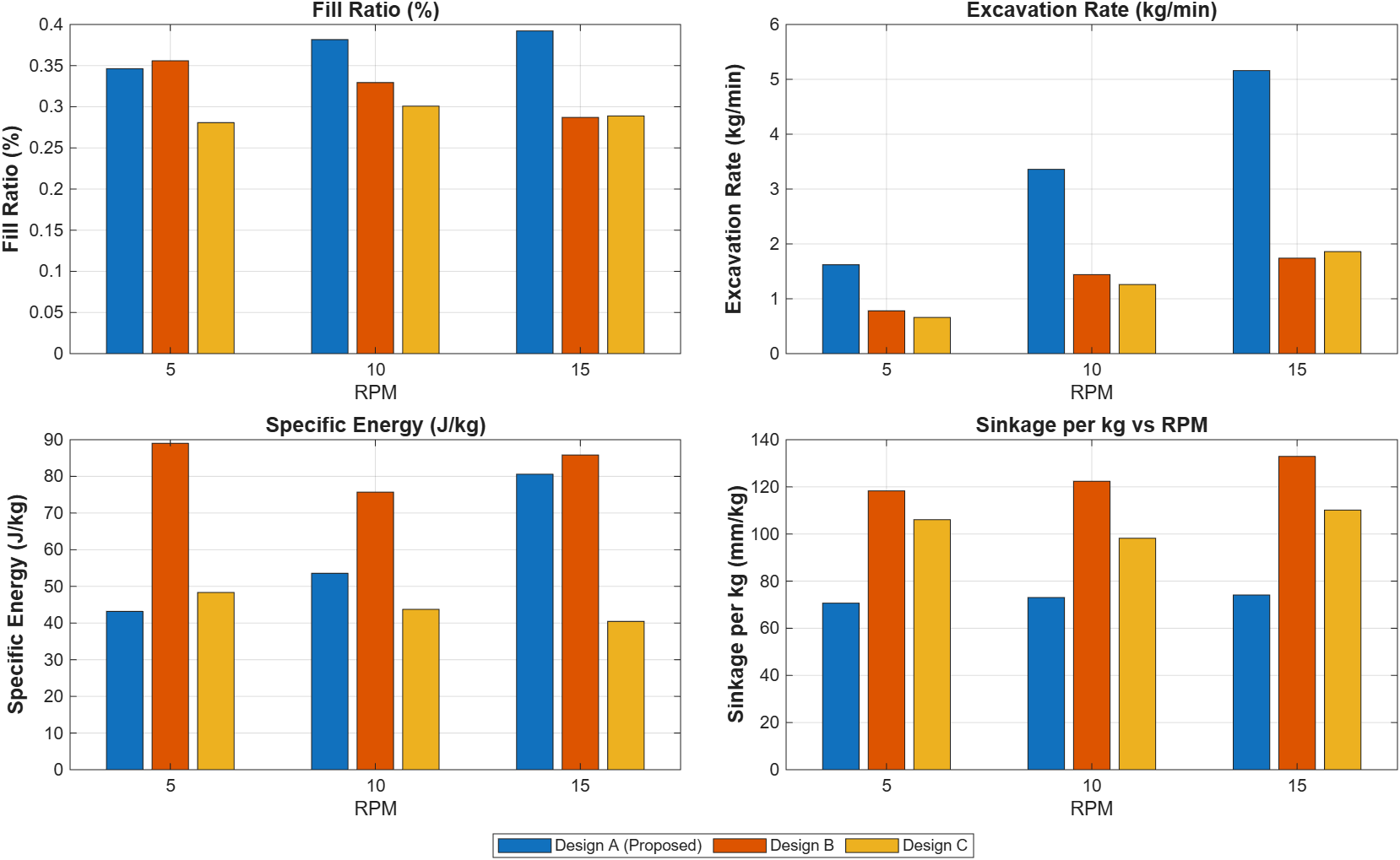}
      \caption{Performance comparison of three wheels tested at 5, 10, and 15 RPM : (a) fill ratio, (b) excavation rate, (c) specific energy, and (d) sinkage per unit mass. The proposed Design A shows superior soil retention and throughput with favorable efficiency compared to Designs B and C}
      \label{Performance}
\end{figure}

Excavation rate results followed a similar trend. At 5~RPM, Design A collected 1.62~kg/min compared with 0.78~kg/min for Design B and 0.62~kg/min for Design C. At 10~RPM, Design A reached 3.36~kg/min, more than twice the rates of Design B (1.44~kg/min) and Design C (1.26~kg/min). At 15~RPM, Design A achieved 5.16~kg/min, which was significantly higher than both reference designs. These results confirm that the spiral cavity in Design A improves soil retention and directly translates into higher excavation throughput across operating speeds.

\subsection{Energy and Efficiency}
Energy analysis showed that Design A required higher total energy input than Design C, but similar levels to Design B. However, when normalized by excavated mass, the specific energy of Design A averaged 59.1~J/kg, which is 29\% lower than Design B (83.5~J/kg) and moderately higher than Design C (44.2~J/kg). Thus, relative to the bucket-drum-inspired Design B, the proposed wheel is both more productive and more energy-efficient. Compared with the traction-oriented Design C, the new wheel consumes more energy per unit mass but achieves 2--3$\times$ higher throughput, demonstrating a favorable trade-off between productivity and efficiency.  
\begin{figure}[t]
      \centering
      \includegraphics[width=.95\linewidth]{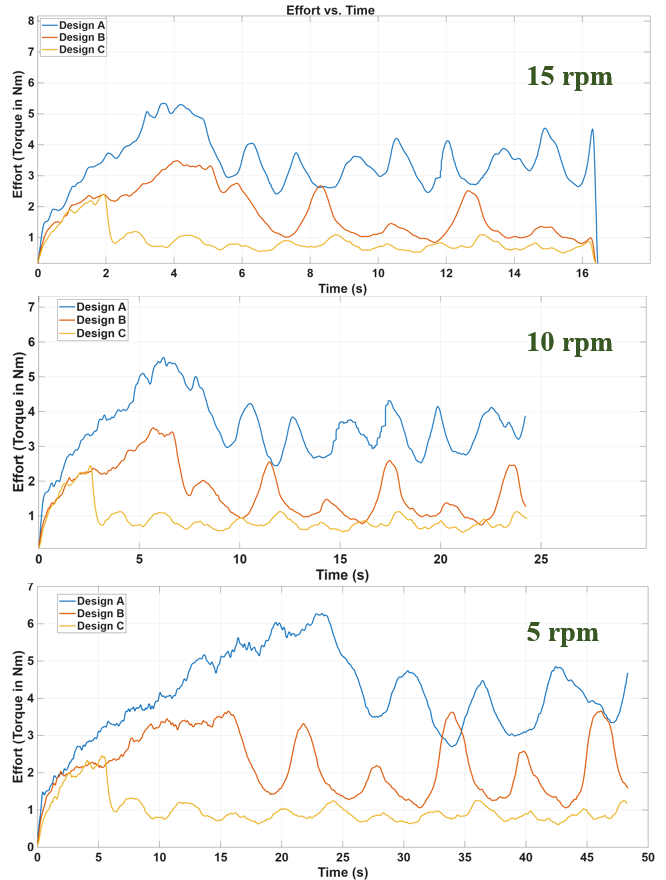}
      \caption{Torque vs. time over four revolutions at 15, 10, and 5 RPM.}
      \label{fig:effort_t}
\end{figure}
\subsection{Sinkage Behavior}
Absolute sinkage measurements indicated that Design A penetrated deeper into the soil than B or C, reflecting its larger excavation volume. Across speeds, Design A sank 93.4, 99.8, and 105.5~mm at 5, 10, and 15~RPM, respectively. By comparison, Design B sank 74.1, 69.7, and 65.1~mm, while Design C penetrated only 54.5, 51.2, and 56.4~mm.To compare terrainability fairly across designs, sinkage was normalized by collected mass. This yielded 72.6~mm/kg for Design A, versus 124.5~mm/kg for Design B and 104.8~mm/kg for Design C. In other words, Design A required 42\% less sinkage per unit mass than B and 31\% less than C, demonstrating superior excavation efficiency relative to soil penetration. Expressed as a fraction of wheel diameter, mean sinkage was 0.36D for A, 0.30D for B, and 0.27D for C.

\subsection{Torque Observations}

Figure~\ref{fig:effort_t} shows actuator torque over time during four-revolution excavation trials at 5, 10, and 15~RPM. These traces offer direct sensor-level insight into wheel–regolith interaction. Baseline torque reflects sustained cutting and transport load, while periodic peaks correspond to cut–carry–dump cycles. The time-integrated signal relates to mechanical energy. Across all speeds, Design A maintains a higher and more consistent torque envelope with repeatable cycles and no stall-like spikes, indicating stable engagement and traction. In contrast, Designs B and C show lower baselines and intermittent peaks, suggesting less consistent material capture and reduced throughput. The steady load profile of Design A aligns with its higher excavated mass and efficiency, while the absence of large excursions implies operation without bogging or instability. These signatures confirm Design A’s superior excavation performance and robustness.

\subsection{Dem Simulation Insight}\
\begin{figure}[thpb]
      \centering
      \includegraphics[width=\linewidth]{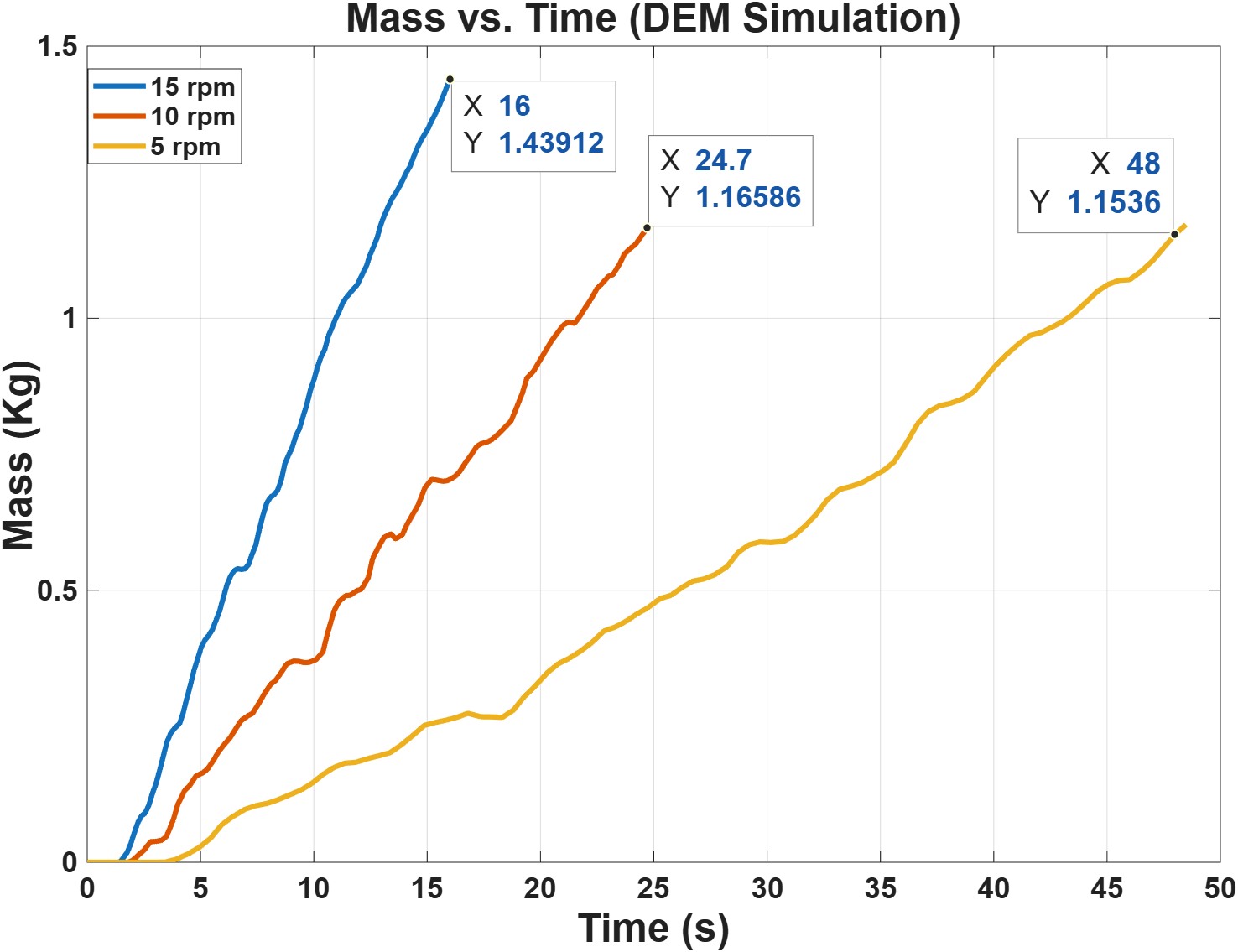}
      \caption{DEM simulation results of Design A showing mass accumulation over time for three rotational speeds: 5, 10, and 15 RPM. The curves indicate that higher rotational speeds yield faster and greater regolith collection, with Design A reaching 1.439~kg at 15~RPM within 16~s. These trends support the observed experimental performance and validate the simulation framework.}
      \label{fig:DEM}
\end{figure}
DEM simulations conducted using Altair EDEM closely matched experimental trends in excavated mass across all tested speeds for Design A. At 15~RPM, the predicted mass (1.439~kg) was within 1.06\% of the measured value, indicating high fidelity. Slightly larger deviations were observed at lower speeds (12.73\% at 5~RPM and 14.65\% at 10~RPM), likely due to increased sensitivity to initial soil packing and boundary effects. Overall, the simulation framework demonstrated strong predictive capability, with a mean absolute error (MAE) of 0.128~kg and mean absolute percentage error (MAPE) of 9.48\%. These results validate the use of DEM for evaluating wheel–soil interaction and support its future use in optimizing fill ratio, traction, and energy metrics.

\begin{table}[t]
  \centering
  \caption{DEM vs.\ experiment on excavated mass per 4 rev (kg).}
  \label{tab:dem_err}
  \setlength{\tabcolsep}{5pt}
  \begin{tabular}{lcccc}
    \toprule
    Condition & Exp.\ (kg) & DEM (kg) & Abs.\ Err.\ (kg) & \% Err. \\
    \midrule
    A @ 5 rpm  & 1.322 & 1.1536  & 0.1684 & 12.73\% \\
    A @ 10 rpm & 1.366 & 1.1659  & 0.2001 & 14.65\% \\
    A @ 15 rpm & 1.424 & 1.4391  & 0.0151 & 1.06\% \\
    \midrule
    \multicolumn{2}{l}{MAE / RMSE / MAPE} &
    \multicolumn{3}{r}{\textbf{0.128} / \textbf{0.140} / \textbf{9.48}\%} \\
    \bottomrule
  \end{tabular}
\end{table}
\subsection{Overall Discussion}
Together, these results demonstrate that the spiral cavity of Design A enhances soil retention and improves excavation efficiency. The soil-holding ratio ($\eta$), introduced as a design index, was reflected in measured fill ratios, validating it as a predictive metric. Although Design A requires higher torque and experiences greater absolute sinkage, it provides significantly more regolith per unit sinkage and per unit energy than the baseline wheels. These findings indicate that integrating a spiral cavity into rover wheels offers a promising direction for ISRU-focused excavation system design.
 
\section{CONCLUSIONS AND FUTURE WORK}

This paper presented the design and experimental evaluation of a spiral-cavity wheel concept for lunar regolith excavation. Unlike conventional open-center wheels, the proposed geometry retains regolith during rotation, resulting in higher fill ratios and excavation rates compared to two baseline wheel designs. Although the spiral-cavity wheel required greater torque and exhibited deeper absolute sinkage, it demonstrated improved regolith collection per unit energy and per unit penetration depth. These results indicate that the proposed geometry offers a favorable trade-off between excavation efficiency and mobility requirements, making it a promising candidate for integrated excavation systems in in-situ resource utilization (ISRU) missions.

Future work will extend this research in several directions. First, additional experiments and Discrete Element Method (DEM) simulations will be conducted across a wider range of soil conditions, including denser and cohesive simulants, to better represent the variability and mechanical behavior of lunar regolith. Second, environmental factors relevant to lunar operations, such as reduced gravity, vacuum conditions, and the abrasive nature of regolith particles, will be considered to evaluate their potential impact on wheel performance, durability, and excavation efficiency.

Third, the retention index $\eta$ introduced in this study will be further developed into a predictive framework capable of estimating excavation performance for different wheel geometries and soil types. Such a model could support the systematic design and optimization of multifunctional excavation wheels for planetary applications. Finally, scalability studies will investigate the applicability of the proposed concept across rover platforms with different sizes and payload capacities, examining the associated torque requirements, cavity depth scaling, and energy consumption.

In parallel with these investigations, future development will include adapting the wheel into a double-bucket configuration that enables simultaneous excavation and mobility, performing horizontal mobility experiments to evaluate traction, slip, and steering behavior, and integrating the wheel into a four-wheeled rover platform for system-level testing. These efforts will advance the concept from controlled laboratory experiments toward multifunctional hardware suitable for robotic excavation and mobility tasks in future lunar ISRU missions.




\addtolength{\textheight}{-12cm}   






\bibliographystyle{IEEEtran}


\bibliography{references}


\end{document}